# PERSISTENT HOMOLOGY MACHINE LEARNING FOR FINGERPRINT CLASSIFICATION

NOAH GIANSIRACUSA, ROBERT GIANSIRACUSA, AND CHUL MOON

ABSTRACT. The fingerprint classification problem is to sort fingerprints into pre-determined groups, such as arch, loop, and whorl. It was asserted in the literature that minutiae points, which are commonly used for fingerprint matching, are not useful for classification. We show that, to the contrary, near state-of-the-art classification accuracy rates can be achieved when applying topological data analysis (TDA) to 3-dimensional point clouds of oriented minutiae points. We also apply TDA to fingerprint ink-roll images, which yields a lower accuracy rate but still shows promise, particularly since the only preprocessing is cropping; moreover, combining the two approaches outperforms each one individually. These methods use supervised learning applied to persistent homology and allow us to explore feature selection on barcodes, an important topic at the interface between TDA and machine learning. We test our classification algorithms on the NIST fingerprint database SD-27.

## 1. INTRODUCTION

Near the end of the 19th century, Sir Francis Galton introduced a systematic framework for fingerprint analysis [Gal92]. One component of his work was to divide all fingerprints into three classes: arch, loop, and whorl. This classification, often with refinements (such as subdividing arches into plain and tented types, and dividing whorls into singles and doubles) is still used by nearly every fingerprint classification scheme today. Automated fingerprint classification is useful for fingerprint matching algorithms, since it reduces the search space involved; it also provides a fertile testing ground for more general explorations in pattern analysis. There is a wide variety of approaches to automated classification, tested on various real and simulated fingerprint databases; see the survey articles [YA04, AM09] and the book chapter [MMJP09, Ch.5] for more background and context, an overview of proposed methodologies, and a comparison of performances.

While there are also many approaches to fingerprint matching, the most widely used ones are based on *minutiae points*, terminations and bifurcations in the ridge pattern [MMJP09, Ch.4.3]. However, there do not appear to be classification algorithms based on minutiae points; in fact, the survey article [YA04, p.80] boldly asserts: "minutiae features are not useful for fingerprint classification." A natural question, then, is whether the global geometric structure of fingerprint classes influences the local geometric structure underlying the distribution of minutiae points in a statistically significant manner—and if so, how to measure this influence. We develop a methodology, based on machine learning and the rapidly growing subject Topological Data Analysis (TDA) [Car09, Zom12, Ghr08, EH10], which shows that classification accuracy rates near the state-of-the-art can be obtained using only minutiae point locations and orientations—though intriguingly, the accuracy rates plummet when only the locations are used.

We test our algorithm on the NIST database SD-27 [NIS00]. This database includes 245 fingerprints for which human experts have identified the minutiae point locations and orientations and determined the fingerprint class. The database also includes JPEG images of ink-roll imprints for each of these fingerprints. We additionally develop a TDA methodology for fingerprint classification that uses only these ink-roll JPEGs, where the only preprocessing performed is simple cropping. The accuracy rate here is not as strong as the minutiae-based approach but nonetheless shows promise; moreover, by combining these two methodologies we obtain an improvement over the approach that only uses minutiae points.

The tool from TDA we rely on is *persistent homology*—using 3-dimensional point clouds in our minutiae-based approach and surface sublevel/superlevel sets in the JPEG-based approach—and we view this paper as establishing the first steps toward a persistent homology framework for fingerprint analysis more generally. One of the fundamental challenges with fingerprint analysis is that distortions occur when fingerprints are imprinted. Even linear distortion such as rotation and rescaling can be problematic, since pre-alignment is difficult when the fingerprint data is noisy and incomplete. But the topological structure of fingerprint data, broadly speaking, is "coordinate-free" and hence independent of rotation and rescaling, thereby obviating the need for pre-alignment.

Non-linear distortions pose a greater challenge and do alter the topology, but by incorporating basic methods from machine learning we can in essence "learn" from the data how such distortions affect the topology. This also provides a concrete setting in which to explore feature selection in persistent homology, a topic that is important for TDA but not yet well-understood [ACC16, Ver16]. By using a larger fingerprint data set and more sophisticated supervised learning algorithms, significant performance improvements should be possible.

**Acknowledgements.** This work began as an undergraduate group research project, led by the first author, at the University of Georgia. We would like to the thank the participants of that group: Joia Bryant, Matthew Burchfield, Bryce Derriso, Kaj Hansen, Tara Kelly, Eric Lybrand, Ashik Nabit, Justin Payan, Irma Stevens, and Sarah Tammen. The first author was partially supported by NSA Young Investigator Grant H98230-16-1-0015 and the third author by NSF IIS-1607919.

## 2. Data

The NIST Special Database 27 (SD-27) is a forensically-oriented fingerprint database designed to help researchers develop and hone matching algorithms for fingerprints of varying quality [NIS00]. It was originally released in 2000 and then re-released in 2010 with higher resolution images. The database comprises 258 fingerprint entries, each containing the following data:

- A *latent* fingerprint image obtained from a crime scene. The finger could be any of the ten digits, and the quality varies from "good" to "bad" to "ugly".
- A police department ink-roll, called a *tenprint*, of the same finger on the same individual as the latent print. While these images are higher quality than even the best latents, they still contain noise, noticeable imperfections, and cropping artifacts inherent to the ink-roll process that pre-dates modern digital fingerprint imaging technology.
- A fingerprint class identified by a human expert: plain arch, tented arch, right slant loop, left slant loop, whorl, or unclassifiable (see Figure 1).
- Four sets of minutiae points that were hand-identified by experts: (1) all the minutiae that were directly discerned on the latent print (called *ideal latent minutiae*), (2) all the minutiae that were directly discerned on the tenprint (called *ideal tenprint minutiae*), (3) the latent minutiae that were identified with corresponding tenprint minutiae (called *matched latent minutiae*), and (4) the tenprint minutiae that were identified with corresponding latent minutiae (called *matched tenprint minutiae*). Each minutiae point is recorded by its coordinates in the fingerprint image; also recorded is the *orientation*, a radial measure of the direction of the bifurcation/termination where the minutiae point occurs (see Figure 2).

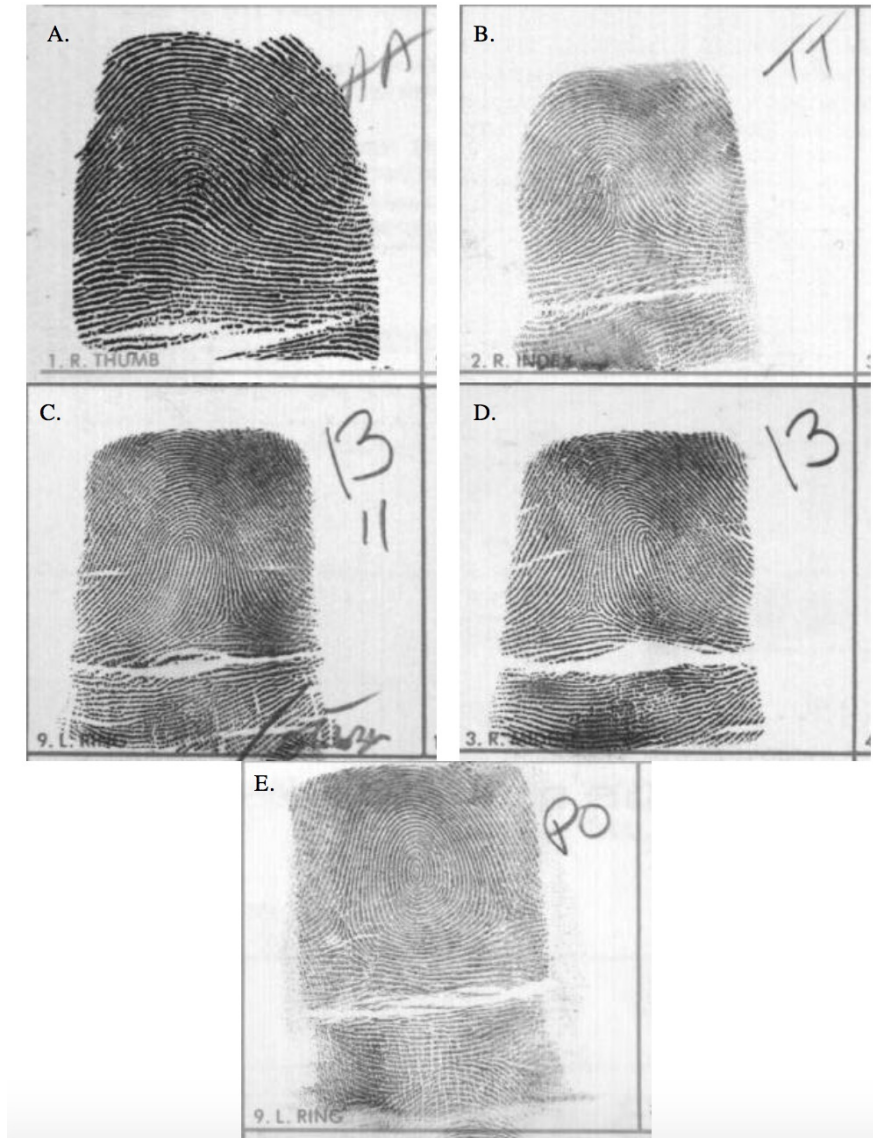

FIGURE 1. Examples of the five fingerprint classes from the data set NIST SD-27: (A) plain arch, (B) tented arch, (C) left slant loop, (D) right slant loop, (E) whorl. In order to have enough instances of each class to train a supervised learning classifier, we consolidate these into three classes: arch (A and B), loop (C and D), whorl (E).

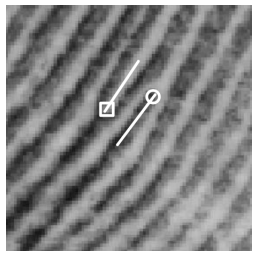

FIGURE 2. Illustration from the NIST SD-27 documentation of the orientation of the two types of minutiae points: bifurcation (square) and termination (circle).

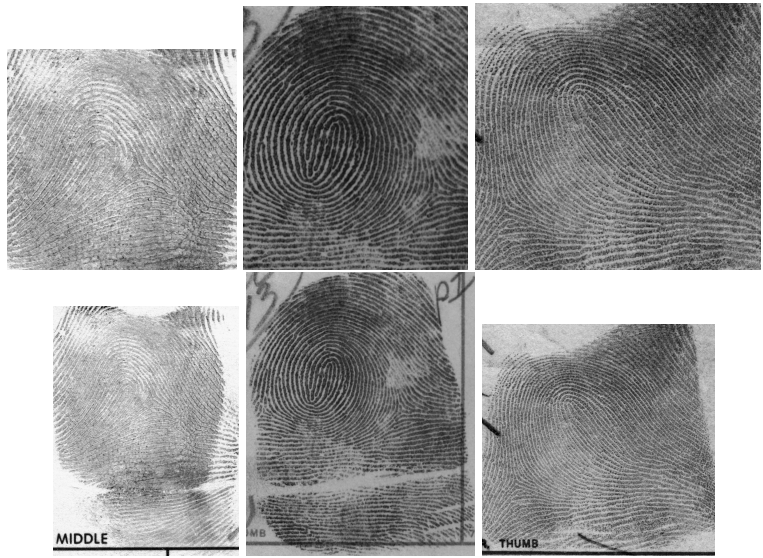

FIGURE 3. Sample illustration of the manual-cropping we performed on the NIST SD-27 fingerprint JPEG images: bottom row are the original NIST images and top row are the cropped images we used in our data analysis.

**Remark 2.1.** In this paper we focus exclusively on the tenprint ink-roll images and the ideal tenprint minutiae, so without further stipulation the terms "ink-roll image" (or "fingerprint JPEG") and "minutiae points" shall refer to these two aspects of the database.

Some fingerprint entries are missing an ink-roll image or have the class listed as unclassifiable; we removed these from the database, resulting in 245 of the original 258 entries. We also reduced the number of classes to Galton's original three (arch, loop, whorl) by merging plain arch with tented arch and right slant loop with left slant loop; this was done to have enough instances of each class to train a supervised learning classifier. In some cases the fingerprint expert could not decide on a single class and so listed multiple possible classes; in such cases we use the first listed class as that is the one in which the expert had the greatest confidence. The distribution of classes in the 245 fingerprints is then: 5.3% arches, 58.4% loops, and 36.3% whorls. According to [WCW93], the naturally occurring probabilities of these classes in a general human population are 6.6% arches, 65.5% loops, and 27.9% whorls, so this NIST database is fairly representative in this regard. The number of minutiae points in these 245 fingerprints ranges from 48 to 193.

As can be seen in Figure 1, the ink-roll images in this database include varying amounts of white space and regions of the finger (often extending past the first joint) and they frequently contain extraneous markings. The only image-processing we perform is manual cropping to mostly eliminate the white space and focus the image on the region of the fingerprint above the main horizontal crease in each finger. Some examples of this cropping are shown in Figure 3.

## 3. METHODOLOGY

One of the main tools in Topological Data Analysis (TDA) is *persistent homology*, which provides a multi-scale measure of the multi-dimensional connectivity of a data set [Zom12, EH10, Ghr08, Car09, BMM$^+$16, CCR13]. We apply persistent homology to the NIST fingerprint data in two distinct ways—using minutiae point data and using ink-roll image data—and then apply machine learning procedures to compute classification accuracy rates.

3.1. **Minutiae-based persistent homology.** The most common format for the data input to persistent homology is a *point cloud*, meaning a finite collection of points in Euclidean space; concretely, for $N$ points in $\mathbb{R}^m$ this means an $N \times m$ matrix listing the coordinates of each point. The points are considered unordered, which means that if the rows of the coordinate matrix are permuted then the persistent homology output is unchanged. In brief, persistent homology replaces each point with a solid sphere and records the topological features of the union of these solid spheres as a function of their radii. The output is a collection of diagrams called *barcodes*: for each dimension $d \geq 0$, the $d$-dimensional barcode is a multiset of intervals, or *bars*, each of whose left endpoint signifies the birth of a $d$-dimensional topological feature and right endpoint signifies its death. The 0-dimensional topological features are connected components and the 1-dimensional topological features are loops. For readers interested in more details and general background, we recommend the above-cited TDA literature, though we attempt to illustrate our methods without assuming prior familiarity with topology.

A natural TDA approach is to view fingerprint minutiae points as a point cloud in $\mathbb{R}^2$—that is, to use minutiae point locations as the input for persistent homology. As we later discuss in the results section, this yields a rather mediocre classification performance. A remarkable improvement is obtained by incorporating the minutiae point orientations. The key insight for how to do this is that persistent homology allows the point cloud to live in any metric space, not just Euclidean space $\mathbb{R}^m$. Minutiae point orientations are simply angles, so they are naturally viewed as points on the unit circle $S^1$. The minutiae points on a given fingerprint then form a point cloud in the manifold $\mathbb{R}^2 \times S^1$, which is a higher-dimensional analogue of the cylinder. There are various natural choices for endowing this product space with a metric. The choices we use are based on the $\ell^1$ metric (also called *taxicab* or *Manhattan*), the $\ell^2$ metric (also called *Euclidean*), and the $\ell^3$ metric. Given the set of $N$ minutiae points

$$\mathbf{p_1} = (x_1, y_1, \theta_1), \ \ldots, \ \mathbf{p_N} = (x_N, y_N, \theta_N) \in \mathbb{R}^2 \times S^1$$

of a fingerprint, we first normalize by replacing each $x_i$ with

$$\frac{x_i - \min_{1 \leq j \leq N}\{x_j\}}{\max_{1 \leq j \leq N}\{x_j\} - \min_{1 \leq j \leq N}\{x_j\}}$$

and similarly for $y_i$, and each $\theta_i$ with $\frac{\theta_i}{\max_{1 \leq j \leq N}\{\theta_j\}}$, so that all coordinates and angles are between 0 and 1. We then define five different metrics computing distances between any pair of these normalized points:

$$d_1(\mathbf{p_i}, \mathbf{p_j}) = |x_i - x_j| + |y_i - y_j| + \theta_{ij}$$

$$d_{1,\frac{1}{3}}(\mathbf{p_i}, \mathbf{p_j}) = \frac{1}{3}(|x_i - x_j| + |y_i - y_j|) + \frac{2}{3}\theta_{ij}$$

$$d_{1,\frac{2}{3}}(\mathbf{p_i}, \mathbf{p_j}) = \frac{2}{3}(|x_i - x_j| + |y_i - y_j|) + \frac{1}{3}\theta_{ij}$$

$$d_2(\mathbf{p_i}, \mathbf{p_j}) = \sqrt{(x_i - x_j)^2 + (y_i - y_j)^2 + \theta_{ij}^2}$$

$$d_3(\mathbf{p_i}, \mathbf{p_j}) = \sqrt[3]{(x_i - x_j)^3 + (y_i - y_j)^3 + \theta_{ij}^3}$$

where

$$\theta_{ij} = \begin{cases} |\theta_i - \theta_j| \text{ if } |\theta_i - \theta_j| \leq \frac{1}{2} \\ 1 - |\theta_i - \theta_j| \text{ if } |\theta_i - \theta_j| > \frac{1}{2} \end{cases}.$$

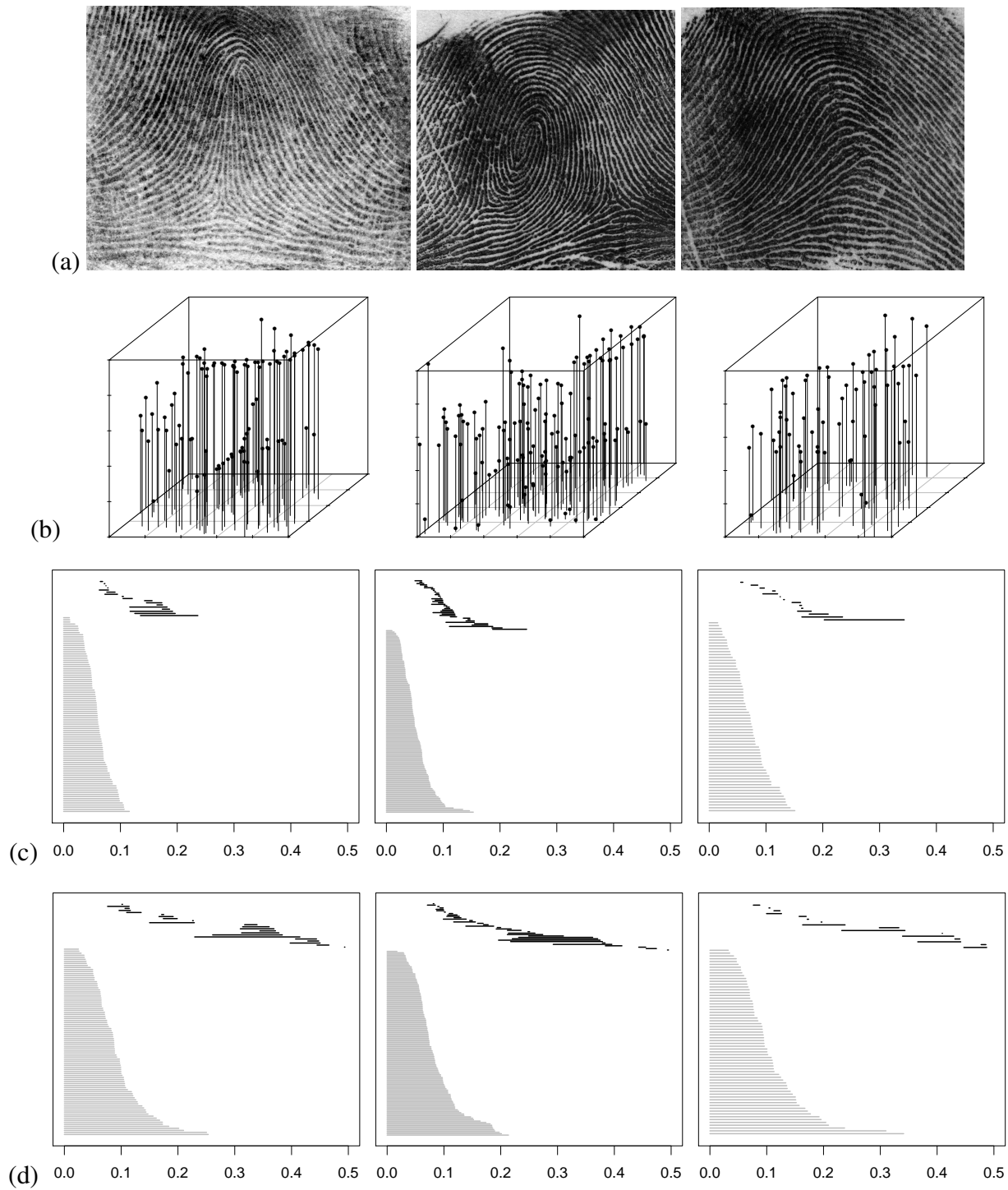

FIGURE 4. (a) Cropped images of a loop (left), whorl (middle), and arch (right). (b) Scatterplots of the corresponding normalized minutiae coordinates (the vertical axis is the orientation, so the top and bottom squares squares should be identified). (c) The 0-dimensional (gray) and 1-dimensional (black) barcodes for the unoriented minutiae point clouds in $\mathbb{R}^2$. (d) The barcodes for the minutiae point clouds in $\mathbb{R}^2 \times S^1$ with the metric $d_2$ defined earlier. Precisely interpreting these barcodes is not necessary; for the purposes of supervised learning we simply need that the barcodes reflect *some* relevant global geometric structure in the fingerprints.

For each of the 245 fingerprints of interest in NIST SD-27, we compute 12 distinct barcodes: the 0- and 1-dimensional persistent homology for the minutiae point clouds in $\mathbb{R}^2$ (with Euclidean metric and orientations ignored) and in $\mathbb{R}^2 \times S^1$ using each of the five different metrics listed above. Persistent homology algorithms have been implemented in a variety of software packages (cf., [OPT$^+$17]); we use the R package called "TDA" [FKLM14]. See Figure 4 for an example.

**Remark 3.1.** There are of course many more (in fact, infinitely many) possible metrics to consider on $\mathbb{R}^2 \times S^1$, but for the purposes of supervised learning one doesn't strive to be comprehensive; instead, one looks for a collection that is small enough to be manageable and not lead to excessive over-fitting while at the same time that discerns enough geometry to be able to separate the fingerprint classes. We settled upon the above choice of five metrics through empirical experimentation.

3.2. **JPEG-based persistent homology.** The point cloud approach to persistent homology is just one of several options. Another form of persistent homology uses a discrete variant of Morse theory. In brief, given a surface in $\mathbb{R}^3$ and a choice of grid resolution, one can compute the topological features of the superlevel sets (or sublevel sets) at height $t$, as a function of $t$, and again assemble these into barcodes. See, e.g., [Zom05, EH08].

We apply this Morse-based method to fingerprint data as follows:

(1) read in the 245 cropped grayscale JPEG fingerprint ink-rolls in NIST SD-27, invert them (so that the background is black instead of white) and store as real-valued matrices;

(2) normalize each matrix by first subtracting off the minimal matrix value and then dividing the new entries by the new maximal value;

(3) use the R package "TDA" [FKLM14] to compute the 0- and 1-dimensional superlevel set persistent homology barcodes of the surface defined by each normalized matrix (see Figure 5, top row), where the grid resolution is provided by the matrix itself (i.e., each grid square is a single matrix entry, which corresponds to a single JPEG pixel).

In principle these barcodes record the global topology of the fingerprint ridge pattern, but there is so much noise and so many minor fluctuations that it is nearly impossible to see this. Regardless, the supervised learning classification pipeline works better when accessing finer geometric information, which we achieve by a novel "slanting" method that we introduce here. After normalizing the JPEG matrices in steps (1) and (2) above, we obtain additional barcodes as follows:

(4) multiply the normalized matrix by the linear function $f(x, y) = y$ then compute the 0- and 1-dimensional barcodes for the superlevel set persistent homology (see Figure 5, middle row); do the same for the function $f(x, y) = x$;

(5) threshold the normalized matrix by setting all values above the mean value of the matrix to 1 and all values below the mean to 0, then multiply by the function $f(x, y) = y$ and compute the 0- and 1-dimensional sublevel set barcodes (see Figure 5, bottom row); do the same for the linear function $f(x, y) = x$ and the non-linear function $f(x, y) = xy$.

This yields 12 barcodes for each fingerprint—six 0-dimensional and six 1-dimensional—just as we have with the minutiae-based approach. The motivation for this slanting process is that the persistent homology of the sublevel/superlevel sets after slanting is related to the "sweep across" persistent homology of the ridge curves and so measures the *tortuosity* of the ridges, not just their global topology (see [BMM$^+$16, ACC16]). The thresholding step accentuates the ridge pattern and provides another way to increase the number of barcodes available (as does alternating between superlevel sets versus sublevel sets).

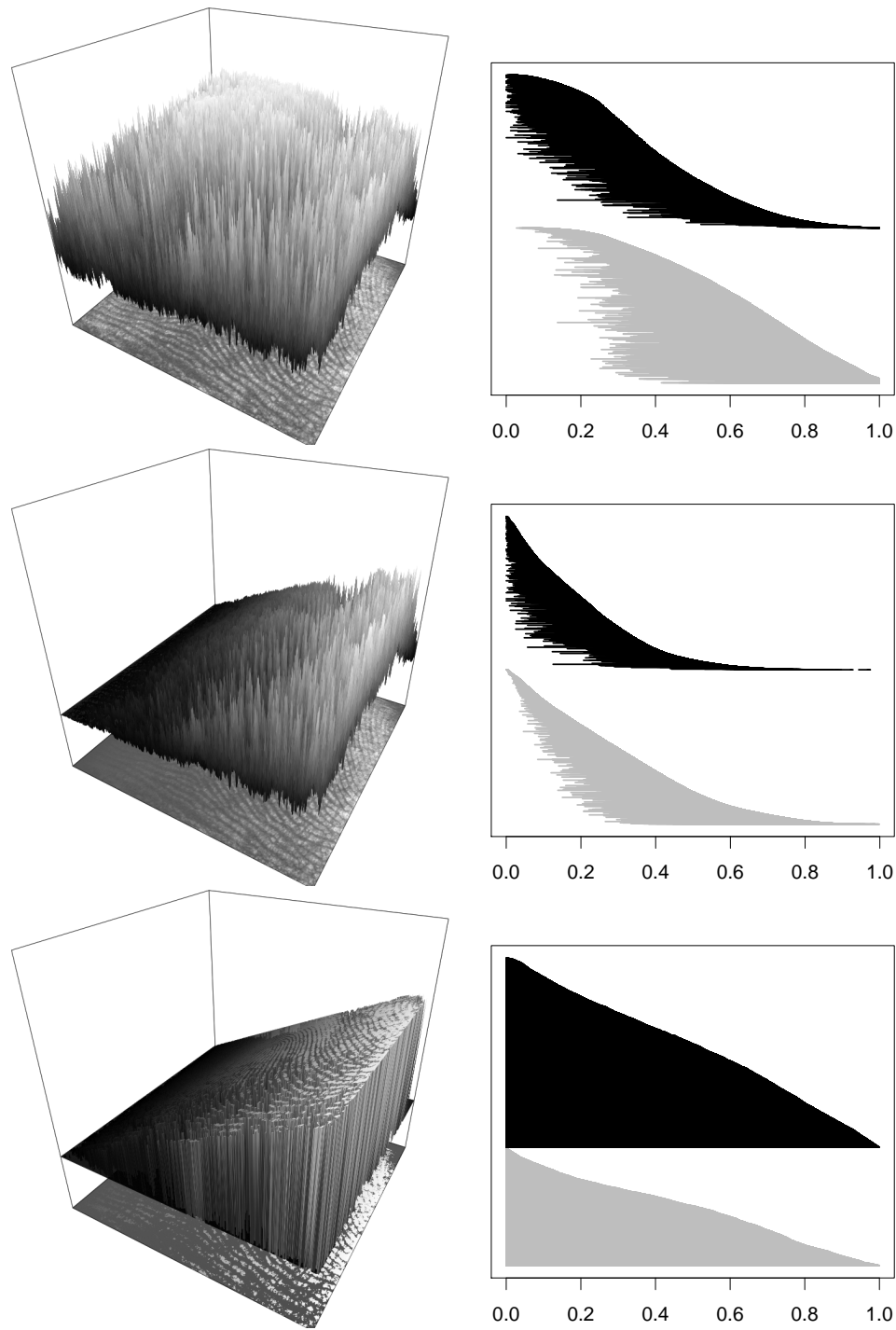

FIGURE 5. Top row: an ink-roll JPEG, after normalizing and inverting, viewed as a 3D surface, and the 0-dimensional (gray) and 1-dimensional (black) barcodes of the superlevel sets. Middle row: the same image after slanting by the function $f(x,y) = y$, and its superlevel set barcodes. Bottom row: same slanting, but first the image is thresholded to convert from grayscale to black-and-white, and here the barcodes use sublevel sets. We employ these variants (and the others described earlier) in an effort to access as much geometry of the ridge pattern as possible.

3.3. **Supervised learning classification.** There are two main ways of combining machine learning algorithms with persistent homology: the *kernel* method where a distance between barcodes must be defined [Bub15, KHN+15], and the *feature* method where a real vector is associated to each barcode, thereby transforming the output of persistent homology into the standard vectorized format to which traditional supervised learning readily applies [ACC16, Ver16]. We utilize the latter method. The art/science of choosing feature vectors for barcodes is still in the early stages and mostly done ad hoc. Consider a $d$-dimensional barcode with $n$ bars, denote the left endpoint of the $i^{\text{th}}$ bar by $x_i$, the right endpoint by $y_i$, and let $y_{max}$ denote the right-most endpoint of any bar appearing in the barcode. We draw our feature vectors from the following collection of real-valued functions defined on the space of barcodes:

- *Polynomial features* [ACC16]. We use the following polynomial features:

$$f_1 = \sum_{i=1}^n (y_i - x_i) \qquad\qquad f_2 = \sum_{i=1}^n n(y_i - x_i)$$
$$f_3 = \sum_{i=1}^n (y_{max} - y_i)(y_i - x_i) \qquad f_4 = \sum_{i=1}^n n(y_{max} - y_i)(y_i - x_i)$$
$$f_5 = \sum_{i=1}^n (y_{max} - y_i)^2(y_i - x_i)^4 \qquad f_6 = \sum_{i=1}^n n(y_{max} - y_i)^2(y_i - x_i)^4.$$

- *Regression coefficients*. First sort the endpoints $y_1, \ldots, y_n$ in decreasing order then use regression to fit a degree $\ell$ polynomial curve to them and interpret the coefficients $c_0^\ell, \ldots, c_\ell^\ell$ as features. (In the literature, this endpoint curve is most commonly used when $d = 0$ and all $x_i = 0$, as in [LCK+11].) We use $\ell = 1$ and $\ell = 2$ in this paper.

- *Statistical features*. The distribution of bars provides the following statistical features:

$$g_1 = \text{mean}\{x_i\},\ g_2 = \text{mean}\{y_i\},\ g_3 = \text{mean}\{y_{max} - y_i\},\ g_4 = \text{mean}\{y_i - x_i\}$$
$$g_5 = \text{median}\{x_i\},\ g_6 = \text{median}\{y_i\},\ g_7 = \text{median}\{y_{max} - y_i\},\ g_8 = \text{median}\{y_i - x_i\}$$
$$g_9 = \text{SD}\{x_i\},\ g_{10} = \text{SD}\{y_i\},\ g_{11} = \text{SD}\{y_{max} - y_i\},\ g_{12} = \text{SD}\{y_i - x_i\}.$$

In total this yields $23 = 6 + 5 + 12$ features for each barcode. Since we have 11 minutiae-based barcodes and 12 JPEG-based barcodes, we obtain $552 = 23(12 + 12)$ features for each fingerprint, though some features will be identically zero so we omit these (and many of these features are highly correlated—a point we return to shortly). As is common practice in machine learning, we normalize the feature vectors so that each has mean zero and standard deviation one.

We use linear discriminant analysis (LDA) classifiers, but since our database is rather small (there are 143 loops, 89 whorls, and only 13 arches) two important steps are necessary: (1) feature selection is first performed to mitigate the curse of dimensionality, and (2) rather than subdividing into training and testing subsets, we use the leave-one-out-cross-validation (LOOCV) method. That is, after fixing an appropriate subset of the 552 features, we consider each fingerprint $F_i$ and train an LCA classifier on the 244 complementary fingerprints $\{F_j\}_{j \neq i}$ then attempt to classify $F_i$; the LOOCV accuracy rate is the number of correct classifications, for $i = 1, \ldots, 245$, divided by 245. This is a standard machine learning technique when dealing with small data sets [JWHT13].

For feature selection, we employ two established techniques. First, once a collection of features of interest has been chosen manually (e.g., all the minutiae-based barcodes or all the JPEG-based barcodes), we use the `findCorrelation` function in the R package "caret" to remove features that are highly correlated with other features, based on a cutoff value. Next, we perform *backwards elimination*; this is the greedy algorithm that removes features one at a time, choosing the single feature at each step whose removal maximizes the LOOCV accuracy rate. As the features are removed in this manner, the accuracy rates generally first increase (as over-fitting is reduced) and then decrease (as under-fitting begins to take over). The peak of this backwards elimination curve indicates a reasonable balance and identifies a moderately sized collection of features that work well together for classifying. Although this tends not to find the optimal subset of features for classifying, it is a reasonable approximation given the computational infeasibility of searching all $2^{552}$ possible subsets.

## 4. Experimental results

Since the dominant fingerprint class in this database is the loop, with 143 out of 245 occurrences, the baseline accuracy rate that all approaches here should be compared to is 58.4%. It is difficult to pin down an accuracy rate for state-of-the-art methods appearing in the literature, since different data sets are used, different preprocessing steps are permitted, different numbers of classes are considered, etc. Useful tables of accuracy comparisons among a wide range of methods are shown on [MMJP09, p.256] and on [YA04, p.90]. Most of these methods use larger databases, which would improve a supervised learning method such as ours, but they also allow four or five classes instead of our three, which certainly makes the classification problem harder. Regardless, we get a coarse estimate by noting that all these published accuracy rates range between 81% and 97%.

It should be noted, however, that some of these reported scores are slightly inflated compared to what we report below by the fact that in some studies fingerprints labelled with multiple classes are considered correctly classified if the algorithm yields any of the listed classes (whereas we only accept the first class listed by the NIST experts); also, some studies allow a certain rejection rate, meaning a fixed percentage of difficult fingerprints are removed from the database prior to computing an overall accuracy rate—we use a 0% rejection rate: every fingerprint in NIST SD-27 that has a class indicated and a matching JPEG image is included.

The accuracy rates we obtain using persistent homology are summarized in Table 1, though first some explanation is in order. For each collection of features named in this table, we choose a cutoff value such that after removing the highly correlated features within this collection determined by `findCorrelation`, there are between 70 and 90 features remaining (except for "unoriented minutiae features," meaning the minutiae point clouds in $\mathbb{R}^2$, since there are only 46 features before removing the highly correlated ones). We then run backwards elimination on these latter features to thin them down further and select the subset of features with the highest accuracy rate among those tested during the backwards elimination. This is the "peak accuracy rate" reported in the table. The "number of features" indicates the size of the subset(s) found by backwards elimination that yields this peak accuracy rate (in some cases multiple subsets achieved the same peak accuracy rate).

TABLE 1. The peak accuracy rates obtained when selecting various sets of features, removing highly correlated ones, then performing backwards elimination on the remaining ones. The rate listed is the maximum obtained this way for each group, and the number of features is the size of the subset(s) of features achieving this rate.

|  | Peak accuracy rate | Number of features |
|---|---|---|
| All 552 features | 93.1% | 32 |
| The 276 0-dimensional features | 82.0% | 19, 25, 28 |
| The 276 1-dimensional features | 93.1% | 32, 33 |
| The 276 minutiae-based features | 91.4% | 48 |
| The 276 JPEG-based features | 77.1% | 37, 40 |
| The 46 unoriented minutiae features | 62.9% | 11 |

Table 2. The confusion matrices for the two classifiers with feature vectors of size 32 that achieved our best accuracy rate, namely 93.1%.

Predicted

|        |       | Loop | Arch | Whorl | Total |
|--------|-------|------|------|-------|-------|
| Actual | Loop  | 138  | 2    | 3     | 143   |
|        | Arch  | 6    | 7    | 0     | 13    |
|        | Whorl | 6    | 0    | 83    | 89    |

32 features from all 552 features

Predicted

|        |       | Loop | Arch | Whorl | Total |
|--------|-------|------|------|-------|-------|
| Actual | Loop  | 137  | 0    | 6     | 143   |
|        | Arch  | 5    | 8    | 0     | 13    |
|        | Whorl | 6    | 0    | 83    | 89    |

32 features from the 1-dimensional ones

The confusion matrices in Table 2 show that for our two best classifiers, nearly half of the arch fingerprints were misclassified as loops but most loops and whorls were correctly classified. A larger training set could help with this issue, but inspecting fingerprints (B) and (C) in Figure 1 shows that loops and arches can in fact appear quite similar.

Table 3. With the notation for our features introduced in §3.3, these are the 32 features selected from among all 552 that achieve our best rate, 93.1%.

| *0-dimensional* | | | | | | | |
|---|---|---|---|---|---|---|---|
| Minutiae cloud $d_3$ metric | $f_5$ | $g_{11}$ | | | | | |
| JPEG surface | $g_{11}$ | | | | | | |
| JPEG $x$-slant | $f_5$ | | | | | | |
| JPEG thresholded $xy$-slant | $g_{11}$ | | | | | | |

| *1-dimensional* | | | | | | | |
|---|---|---|---|---|---|---|---|
| Minutiae cloud $d_1$ metric | $g_8$ | | | | | | |
| Minutiae cloud $d_{1,\frac{1}{3}}$ metric | $f_5$ | $g_8$ | | | | | |
| Minutiae cloud $d_{1,\frac{2}{3}}$ metric | $f_3$ | $f_6$ | | | | | |
| Minutiae cloud $d_2$ metric | $c_1^1$ | $g_1$ | $g_3$ | $g_4$ | $g_8$ | $g_{10}$ | $g_{12}$ |
| Minutiae cloud $d_3$ metric | $c_2^2$ | $f_6$ | $g_3$ | | | | |
| JPEG surface | $f_5$ | | | | | | |
| JPEG $y$-slant | $g_3$ | | | | | | |
| JPEG thresholded $x$-slant | $f_2$ | $g_{11}$ | | | | | |
| JPEG thresholded $y$-slant | $c_1^2$ | $c_2^2$ | $g_4$ | $g_{11}$ | $g_{12}$ | | |
| JPEG thresholded $xy$-slant | $c_2^2$ | $g_1$ | $g_2$ | | | | |

We see from Table 3 that for the best subset of features we found among the 552, most but not all are 1-dimensional, there is a mix of minutiae-based and JPEG-based, and certain functions (such as $g_{11} = \text{SD}\{y_{max} - y_i\}$) show up much more frequently than others.

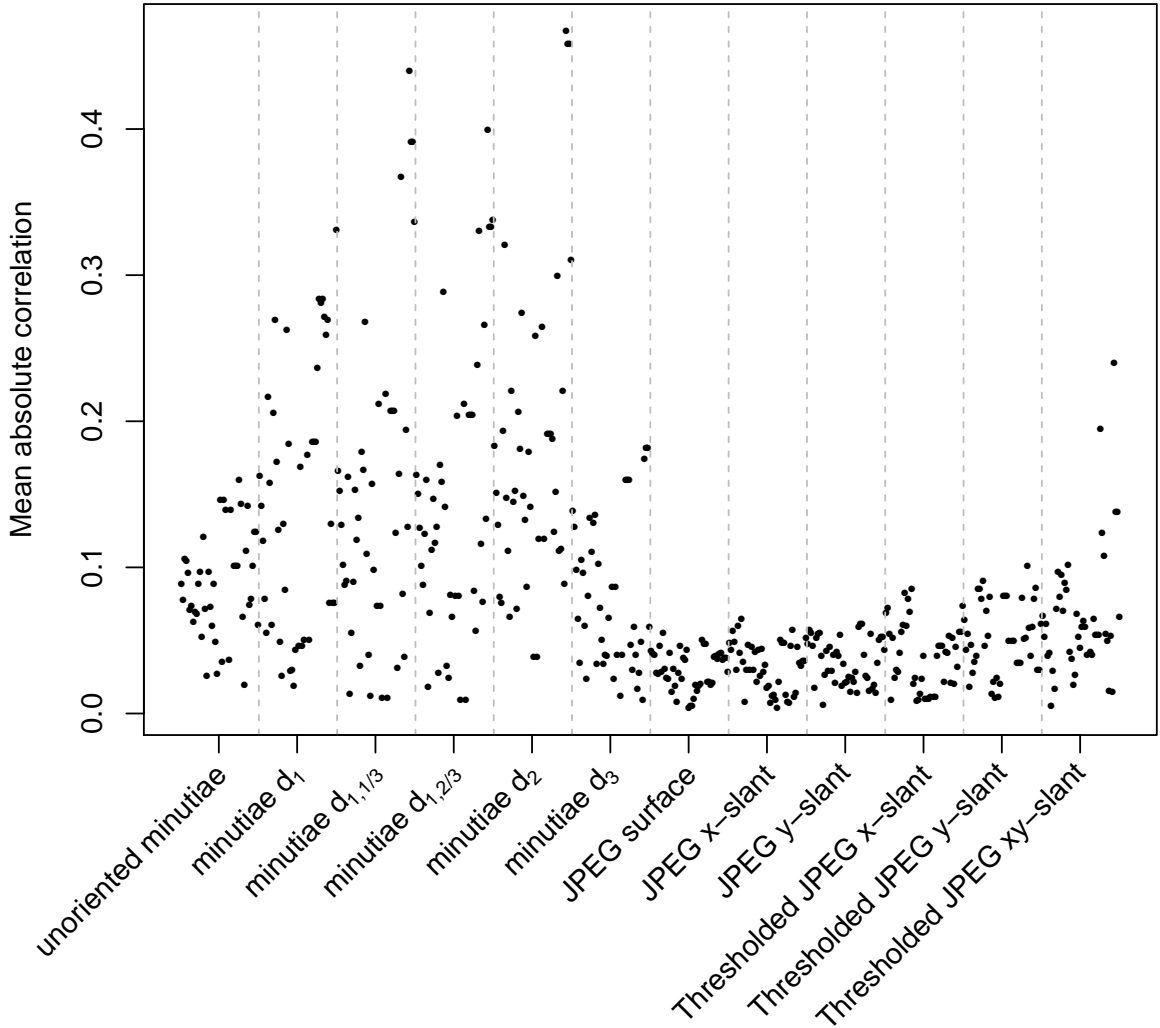

FIGURE 6. An approximation to the classification strength of each of the 552 features. For each of the three 2-class problems, we code the class as 0 or 1 and compute the absolute value of the correlation between each feature and this coded class and then plot the mean of these three values. The features are grouped according to the type of barcode on which they are computed. While correlation only provides a rough approximation to the classification strength of each feature (and completely ignores the complex interplay between the features during classification) many intriguing trends are nonetheless revealed by this plot.

To get a rough sense of the ability of each feature to classify fingerprints, we can consider the three 2-class problems (arch/loop versus whorl, arch/whorl versus loop, and loop/whorl versus arch) and for each one we code the two classes as 0 and 1 then compute the absolute value of the correlation between each feature and the coded class. In Figure 6 we plot the mean of these three absolute-value class correlations for all 552 features; in Figure 7 we scatterplot the three correlation values for the 276 minutiae-based features (left) and for the 276 JPEG-based features (right), indicating which features are 0-dimensional and which are 1-dimensional. Several trends are apparent from these plots:

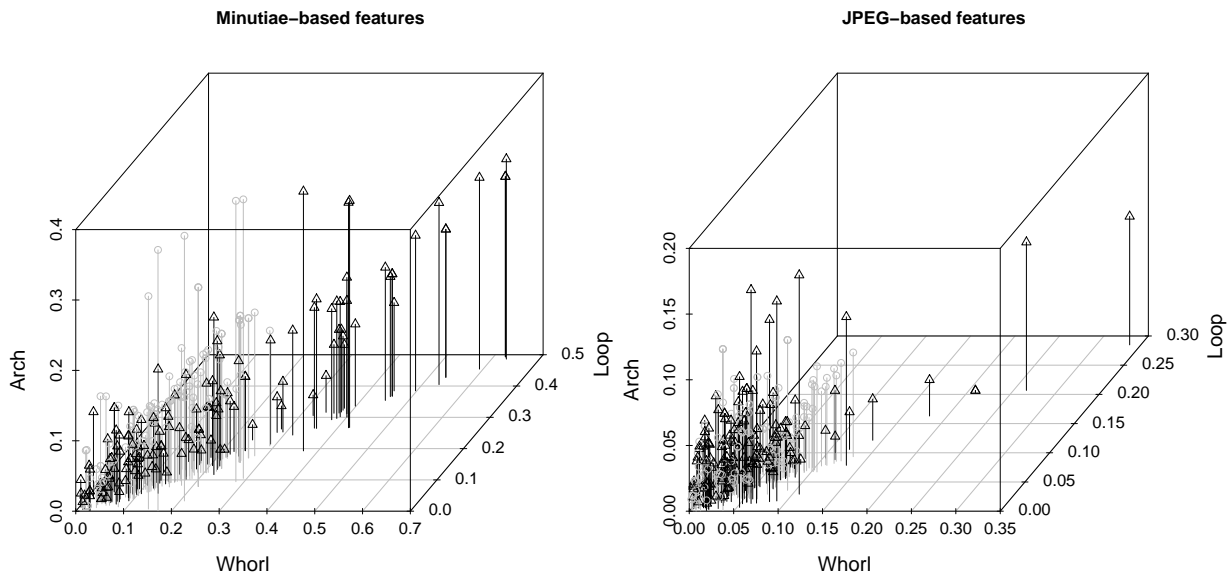

FIGURE 7. Scatterplots estimating the ability of the 276 minutiae-based features (left) and the 276 JPEG-based features (right) to separate out each of the three fingerprint classes. For each of the three 2-class problems we code the class as 0 or 1 then compute the absolute value of the correlation between each feature and the class. The $x$-axis is the absolute correlation for separating out whorls, the $y$-axis is for loops, and the $z$-axis is for arches; the gray circles are 0-dimensional features and the black triangles are 1-dimensional features.

- The minutiae-based features are mostly much stronger than the JPEG-based features, except a few of the thresholded JPEG $xy$-slant features are fairly competitive.
- The unoriented minutiae-based features are not as strong as any of the oriented variants.
- For each metric on the oriented minutiae-based features, there are some very strong features and some very weak features.
- The strongest features tend to be 1-dimensional, though the 0-dimensional features provide some crucial arch versus loop/whorl separation.
- For the most part, features that are good at separating out loops are also good at separating out whorls, and vice versa; on the other hand, some features (mostly 0-dimensional for the minutiae-based and 1-dimensional for the JPEG-based) are much better at separating out arches than they are at the other two classes.

## 5. CONCLUSION AND DISCUSSION

The assertion in [YA04] that minutiae points are not useful for classification is unfounded and, evidently, untrue. While it is difficult to compare the performance of automated fingerprint classification algorithms across different databases and methodologies, our minutiae-based persistent homology (with a peak accuracy rate of 91.4% in our 3-class setting) appears to perform squarely within the range of accuracies demonstrated by other published fingerprint classification methods. Interestingly, however, the performance precipitously drops when we use only the minutiae locations, rather than the locations and orientations (indeed, our peak accuracy rate there is 62.9%, barely above the baseline comparison rate of 58.4% achieved by always guessing "loop"). From a practical perspective, this reduces the utility of our method since many databases do not include the minutiae orientations and automatically extracting them is an additional non-trivial layer of pre-processing. From a theoretical perspective, however, this prominent role played by the minutiae orientations provides some novel insight into how the global geometry of fingerprint classes influences the local geometry of the ridge pattern—though making this precise remains a challenge.

Turning to our JPEG-based persistent homology approach, we find a peak accuracy rate, 77.1%, that is significantly above the baseline comparison but somewhat short of the state-of-the-art methods in the literature. However, the fact that the only preprocessing this method uses is cropping (not rotating, translating, smoothing, cleaning the image, etc.) means that it still shows significant promise and should be studied further. Moreover, by combining the JPEG-based approach with the minutiae-based approach, our peak accuracy rate increases by 1.7 percentage points (from 91.4% to 93.1%), a small but non-trivial improvement. Intriguingly, in the set of features selected by backwards elimination to achieve this peak accuracy rate, nearly half are JPEG-based (15 out of 32, see Table 3) even though in terms of classification (Table 1) and correlation (Figure 6) the minutiae-based features are much stronger than the JPEG-based ones. One possible explanation for this is that the minutiae-based features classify quite well and do the bulk of the work, but to push the rate even higher one needs to add geometric information that is dissimilar to what the minutiae data discern—and while the JPEG-based features accomplish this, each one is so weak that it takes a fairly large number of them to yield a noticeable improvement.

Since directly inferring geometric structure from complicated barcodes is not plausible, we based our persistent homology approach on supervised learning; consequently, our results are strongly influenced by the size of the data set. We wanted to use both oriented minutiae points and matching JPEG ink-roll images and this narrowed our choice of data set down to NIST SD-27, which is rather small compared to most fingerprint databases used in the fingerprint community for training and testing purposes. In particular, we only had 13 arches, which makes "learning" the shape of their barcodes quite challenging (and indeed the confusion matrices in Table 2 reveal that arches have a vastly greater frequency of misclassification than the other two classes).

Persistent homology is essentially invariant under translation and rotation—except that our JPEG method involves the choice of orthogonal directions to slant the surface, but we do not believe the results would be very sensitive to small changes in these directions—so there is little challenge to using our methodology across data sets, and no challenge at all to doing so for the minutiae-based approach, and doing so should significantly improve accuracy rates. Practically speaking, we believe that one could, for instance, train a minutiae-based persistent homology classifier on large data sets and then apply it to any new data set that contains minutiae point locations and orientations, no matter how they are recorded.

More generally, we believe that persistent homology provides an important new tool for attacking difficult pattern analysis problems such as fingerprint classification. While much remains to be understood regarding the interface between persistent homology and machine learning, we hope that this study helps provide some insight into feature selection on barcodes—both by summarizing and introducing a convenient collection of features (§3.3) and by exploring their impact on a concrete, well-studied classification problem.

We close by suggesting some topics for future work: (1) training and testing the method introduced in this paper on a larger database, or across databases, to see how much the accuracy can be improved; (2) incorporating more sophisticated techniques from machine learning, since there is likely room here as well for significant improvement; (3) adapting and applying persistent homology methods to other problems in fingerprint analysis, such as matching noisy latent prints.